\documentclass{article}
\usepackage{ijcai22}

\usepackage{times}
\usepackage{soul}
\usepackage{url}
\usepackage[hidelinks]{hyperref}
\usepackage[utf8]{inputenc}
\usepackage[small]{caption}
\usepackage{graphicx}
\usepackage{amsmath}
\usepackage{amsthm}
\usepackage{booktabs}
\usepackage{algorithm}
\usepackage{algorithmic}
\usepackage{natbib}
\usepackage{amssymb}
\usepackage{amsfonts}
\usepackage{mathtools}
\usepackage{subfigure}
\usepackage{caption}

\title{Multimodal Sequential Generative Models for \\ Semi-Supervised Language Instruction Following}

\author{
Kei Akuzawa$^1$
\and
Yusuke Iwasawa$^1$\and
Yutaka Matsuo$^1$
\affiliations
$^1$The University of Tokyo, Japan\\
\emails
\{akuzawa-kei,iwasawa,matsuo\}@weblab.t.u-tokyo.ac.jp
}

\begin{document}

\maketitle

\begin{abstract}
  Agents that can follow language instructions are expected to be useful in a variety of situations such as navigation.
  However, training neural network-based agents requires numerous paired trajectories and languages.
  This paper proposes using multimodal generative models for semi-supervised learning in the instruction following tasks.
  The models learn a shared representation of the paired data, and enable semi-supervised learning by reconstructing unpaired data through the representation.
  Key challenges in applying the models to sequence-to-sequence tasks including instruction following are learning a shared representation of variable-length mulitimodal data and incorporating attention mechanisms.
  To address the problems, this paper proposes a novel network architecture to absorb the difference in the sequence lengths of the multimodal data.
  In addition, to further improve the performance, this paper shows how to incorporate the generative model-based approach with an existing semi-supervised method called a speaker-follower model, and proposes a regularization term that improves inference using unpaired trajectories.
  Experiments on BabyAI and Room-to-Room (R2R) environments show that the proposed method improves the performance of instruction following by leveraging unpaired data, and improves the performance of the speaker-follower model by 2\% to 4\% in R2R.
\end{abstract}

\section{Introduction}

Agents that can follow language instructions are expected to be useful in a variety of situations and a long-standing topic of interest in robotics \citep{michael2011pancake}.
Various simulation environments, such as BabyAI \citep{chevalier-boisvert2018babyai} and Room-to-Room (R2R) \citep{anderson2018vision} have been proposed as benchmarks.
The combination of deep neural networks (DNNs) with imitation learning (IL) or reinforcement learning (RL) has been shown to enable navigation in realistic 3D environments \citep{hong2021vln} and the execution of complex instructions consisting of multiple subtasks \citep{chevalier-boisvert2018babyai}.
However, numerous paired trajectory and language data is required for training the agent parameterized by DNNs, e.g., hundreds of thousands even on a gridworld \citep{chevalier-boisvert2018babyai}.

Since the amount of paired data is typically limited, \citet{daniel2018speakerfollower} proposed a semi-supervised learning method called the speaker-follower model.
This method uses a small amount of paired data to learn a mapping from trajectory to language, called a speaker.
Then, the trained speaker generates annotations for trajectory-only data (referred to as unpaired trajectory data) to create pseudo-paired data.
Finally, a policy (follower) is trained with IL using the pseudo-paired data.
Currently, the speaker-follower model is a dominant approach to utilize unpaired trajectory data, and many subsequent studies (e.g., \citet{tan2019envdrop,yu2020take}) have been proposed.
However, although unpaired data is assumed to be available, the speaker must be trained with only paired data in these methods.

Different from these studies, this paper investigates a generative model-based approach.
We first introduce a generative model-based formulation for semi-supervised learning iin instruction following by applying a multimodal variational autoencoder (M-VAE) \citep{wu2018multimodal} to the task.
The approach enables semi-supervised learning by learning a shared representation of the two modalities (trajectories and languages) and reconstructing the unpaired data through the representation.
Moreover, since the existing M-VAEs are not suited for sequence-to-sequence (seq2seq) tasks including instruction following, we extend M-VAE and propose a multi-modal sequential variational autoencoder (MS-VAE).
Specifically, since the follower and speaker perform seq2seq conversion, the attention mechanism \citep{luong2015effective} plays a significant role and was often employed in previous studies \citep{anderson2018vision,chevalier-boisvert2018babyai}.
However, previous studies on M-VAE \citep{wu2018multimodal,shi2019variational} left the incorporation of attention as an open problem.
We assume that the existing M-VAEs were incompatible with the attention because the M-VAEs model the shared representation of variable-length multimodal data using \textit{a single vector}, but the attention needs \textit{a sequential structure} in the latent representation.
On the other hand, MS-VAE can be equipped with the attention because it samples \textit{fixed-length sequential latent variables} while absorbs the difference in the sequence lengths of two modalities using a novel module called bottleneck attention.

Also, to further improve the performance, this paper shows how to incorporate MS-VAE with the existing method, speaker-follower model, and proposes a regularization term that improves inference using unpaired trajectories.
Specifically, by performing the cross-modal conversion, i.e., converting languages into trajectories or vice versa, MS-VAE can be used as both a follower and speaker.
Therefore, MS-VAE can utilize unpaired trajectory and language data for improving both the follower and speaker.
The improved follower will directly facilitate the instruction following task; moreover, the speaker can also be used in a way complementary to the speaker-follower model.
Also, as shown in our experiments, the speaker tends to overfit when trained with the reconstruction of unpaired trajectories.
Then, we propose the regularization which prevents overfitting by matching the embedding distributions of the unpaired and paired trajectories.

In the experiments using BabyAI and R2R, we confirmed that MS-VAE contributes to instruction following in two ways.
First, compared to the follower trained with only paired data, MS-VAE used as a follower achieved a higher success rate of tasks.
Second, the incorporation of MS-VAE as a speaker into the speaker-follower model improves performance.
In particular, it improves the success rate of the speaker-follower model by 2\% to 4\% in R2R.

The contributions can be summarized as follows.
\begin{itemize}
    \item We provide a generative model formulation of instruction following.
          This formulation helps to introduce techniques from the field of generative models (e.g., semi-supervised learning) into instruction following.
    \item We propose a novel neural network module,
          which helps multimodal generative models to learn the shared representation of variable-length multimodal data and to incorporate the attention mechanism.
    \item To further improve the performance of semi-supervised instruction following,
          we show how to incorporate the proposed method with the speaker-follower model and propose a novel regularization term.
\end{itemize}

\section{Preliminary: Semi-Supervised Learning in Instruction Following}

\subsection{Problem Settings} \label{sec:preliminary:setting}

The goal of instruction following is to obtain a policy that can execute language instructions $y_i = [y_{i, 1}, ..., y_{i, L_i}]$.
The instruction specifies the task, e.g., \textit{``Go down the stairs''}.
The policy outputs an action $a_{i, t}$ given $y_i$ and observations $[o_{i, 1}, ..., o_{i, t}]$ up to timestep $t$.
In training the policy, following \citet{daniel2018speakerfollower}, this study focuses on semi-supervised IL settings: paired dataset $D_{p} = \{\tau_i, y_i\}_{i=1}^M$ and unpaired trajectory dataset $D_{u} = \{\tau_i\}_{i=1}^N$ are assumed to be available.
Here, $M$ and $N$ are the sample sizes.
$\tau_i = [o_i, a_i]$ is the trajectory that contains actions $a_i = [a_{i, 1}, ..., a_{i, T_i}]$ and observations $[o_{i, 1}, ..., o_{i, T_i}]$.
$L_i$ and $T_i$ are the lengths of the language $y_i$ and trajectory $\tau_i$, which are different for each episode $i$ (the subscript $i$ will be dropped where no ambiguity results).

\subsection{Baseline Follower Architecture}

Here we provide a brief overview of the baseline follower architecture used in R2R \citep{anderson2018vision} and BabyI \citep{chevalier-boisvert2018babyai} because the architecture of MS-VAE is based on it.
The follower adapted the standard attention-based seq2seq model \citep{luong2015effective} to instruction following.
Specifically, it encodes the language $y=[y_1, . , y_{L}]$ with an encoder $f_{enc}$ (e.g., LSTM) and obtain the hidden state $[h_1, ..., h_L] = f_{enc} (y)$.
The decoder $f_{dec}$ (e.g., LSTM) then calculates the hidden state $h^\prime_t = f_{dec}(a_{1:t-1}, o_{1:t})$ using the past observations and actions.
Subsequently, the attention $f_{att}$ calculates the context $c_t = f_{att} (h^\prime_t, h)$, using $h^\prime_t$ as queries and $h = [h_1, . , h_L]$ as keys and values.
Finally, the context $c_t$ and $h^\prime_t$ is used to generate the action $a_t$ with an action predictor $a_t = f_{act} (c_t, h^\prime_t)$.

\subsection{Existing Method: Speaker-Follower Model} \label{sec:preliminary:speaker-follower}

The speaker-followr model performs data augmentation utilizing a speaker.
The speaker is also an attention-based seq2seq model that maps $\tau$ into $y$.
In their method, the speaker is trained with supervised learning using the paired datasets.
Then, the speaker generates languages $\{\hat{y}_i\}_{i=1}^N$ for the unpaired trajectories $D_{u} = \{\tau_i\}_{i=1}^N $.
Finally, the pseudo paired data $\{\tau_i, \hat{y}_i\}_{i=1}^N$ is used to train the follower.

\section{Proposed Method}

\subsection{Generative Model-based Formulation} \label{sec:prop:pgm}

Here we provide a generative model-based formulation for instruction following.
First, we define the generative model and obtain its objective function by applying M-VAE \citep{wu2018multimodal} to the instruction following settings.
Then, we explain how the generative model can be used as the follower and speaker, and how to incorporate it with the existing method, speaker-follower model.

Specifically, we consider a multimodal generative model $p_\theta(z, y, a | o) = p_\theta(z)  p_\theta(y|z) p_\theta(a|z, o)$.
Note that the generative model can represent a standard M-VAE formulation, $p_\theta(z, x_{(1)}, x_{(2)})$, when we substitute $x_{(1)} = a | o$ and $x_{(2)} = y$.
With this generating process, $z$ is expected to become a factor common to $x_{(1)}=a|o$ and $x_{(2)}=y$, i.e., $z$ will be a shared representation with information about the task of the agent.
Also, since the true posterior $p_\theta(z| x_{(1)}, x_{(2)})$ is unknown, it is approximated by the variational posterior $q_\phi(z | x_{(1)}, x_{(2)}) = q_\phi(z| y, a, o)$ parameterized by DNNs $\phi$.
Specific designs for the variational posterior have been proposed by several studies on M-VAE (e.g., \citet{suzuki2016joint,wu2018multimodal}).
This study uses the posterior defined by mixture of experts:
\begin{eqnarray}
    q_\phi(z | y, a, o) = \frac{1}{2} q_\phi(z|y) + \frac{1}{2} q_\phi(z|a, o), \label{eq:moe}
\end{eqnarray}
which was shown by \citet{shi2019variational} to be better at cross-modal conversion than prior methods.

Next, we describe the objective functions.
When using paired data $D_p$, we maximize the following evidence lower bound (ELBO) of the marginal likelihood $p_\theta(y, a|o)$:
\begin{eqnarray}
    \mathcal{J} =& \frac{1}{2} \mathbb{E}_{q_\phi(z|y)} [\log \frac{p_\theta(z) p_\theta(y|z) p_\theta(a|z, o) }{q_\phi(z|y, a, o)} ] \nonumber \\
                &+ \frac{1}{2} \mathbb{E}_{q_\phi(z|a, o)} [\log \frac{p_\theta(z) p_\theta(y|z) p_\theta(a|z, o) }{q_\phi(z|y, a, o)} ]. \label{eq:elbo-mvaelf}
\end{eqnarray}
Because Eq. \ref{eq:elbo-mvaelf} is intractable for analytical computations, we use its lower bound $\bar{\mathcal{J}}$ for optimization (see, Appendix A).

Also, when using unpaired trajectories $D_u$, we use the marginal likelihood $p_\theta(a|o)$ of the generative model $p(z, a| o)=p_\theta(z)p_\theta(a|z, o)$.
Specifically, we maximize the following ELBO of $p_\theta(a|o)$:
\begin{eqnarray}
    \mathcal{V}  = \mathbb{E}_{q_\phi(z|a, o)} [\log p_\theta(a|z, o)] + D_{\mathrm{KL}} [q_\phi(z|a, o)|p_\theta(z)] \label{eq:traj-vae}
\end{eqnarray}
This objective function allows us to train the trajectory encoder $q_\phi(z|a, o)$ and action decoder $p_\theta(a|z, o)$ using unpaired trajectory data.

As noted in Section \ref{sec:preliminary:setting}, both paired and unpaired data are available for optimization.
Thus, Eq. \ref{eq:elbo-mvaelf} is used for training with paired data, and Eq. \ref{eq:traj-vae} is used for training with unpaired trajectory data.
Therefore, the objective function is the weighted sum of the following terms:.
\begin{eqnarray}
    \max \mathbb{E}_{\{\tau, y\} \in D_{p}} [\bar{\mathcal{J}}] + \gamma \mathbb{E}_{\tau \in D_{u}} [\mathcal{V}]. \label{eq:elbo-sum}
\end{eqnarray}
Here, $\gamma$ is the weighting parameter.
In practice, the optimization is performed with mini-batches.

Next, we explain how to use these generative and inference models in instruction following.
First, an agent usually outputs actions one by one rather than simultaneously outputting a whole action sequence.
To represent it, we introduce a timestep $t$ into the action decoder $p_\theta (a|z, o)$ to enable the successive selection of actions.
Specifically, we define $p_\theta (a|z, o)$ as follows.
\begin{eqnarray}
    p_\theta (a|z, o) = \Pi_{t=1}^T p_\theta(a_t|z, o_{1:t}, a_{1:t-1}) \label{eq:policy}
\end{eqnarray}
Eq. \ref{eq:policy} means the decoder $p_\theta (a|z, o)$ is defined as an autoregressive model.
$p_\theta(a_t|z, o_{1:t}, a_{1:t-1})$ indicates that action $a_t$ is determined using the past observation $o_{1:t}$, the past action $a_{1:t-1}$, and the latent representation $z$.
Also, the language decoder $p_\theta(y|z)$ is defined by an autoregressive model, which is commonly used for language modeling, as follows:
\begin{eqnarray}
    p_\theta (y|z) = \Pi_{t=1}^L p_\theta(y_t|z, y_{1:t-1}).
\end{eqnarray}

With these decoders, the model can be used as a follower or speaker by performing a cross-modal conversion.
When used as a follower, the latent representation $z$ of the task is first sampled using the language encoder $q_\phi(z|y)$.
Subsequently, the action decoder (Eq. \ref{eq:policy}) selects actions as follows.
\begin{eqnarray}
    a_t \sim p_\theta(a_t|z, o_{1:t}, a_{1:t-1}), \; \mathrm{where} \; z \sim q_\phi (z|y)  \label{eq:rollout}
\end{eqnarray}
In practice, we can replace $z$ with mean for $q_\phi (z|y)$.
By contrast, when the model is used as a speaker, $q_\phi(z|a, o)$ and $p_\theta(y| z)$ are regarded as the encoder and decoder, respectively.
Using them, seq2seq conversion is performed in a manner, similar to Eq. \ref{eq:rollout} as follows.
\begin{eqnarray}
    y_t \sim p_\theta(y_t|z, y_{1:t-1}), \; \mathrm{where} \; z \sim q_\phi (z|a, o). \label{eq:rollout-lang}
\end{eqnarray}

The generative model can be used for instruction following in two ways.
The first approach (denoted as MS-VAE-follower) uses a trained model as a follower.
The second approach (MS-VAE-speaker-follower) uses a trained model as a speaker and incorporates it into the speaker-follower model (See, Section \ref{sec:preliminary:speaker-follower}).
These two approaches are evaluated in Section \ref{sec:exp:two-approach}.

\subsection{Bottleneck Attention Module} \label{sec:prop:attn}

\begin{figure}[t]
  \centering
  \includegraphics[width=\columnwidth]{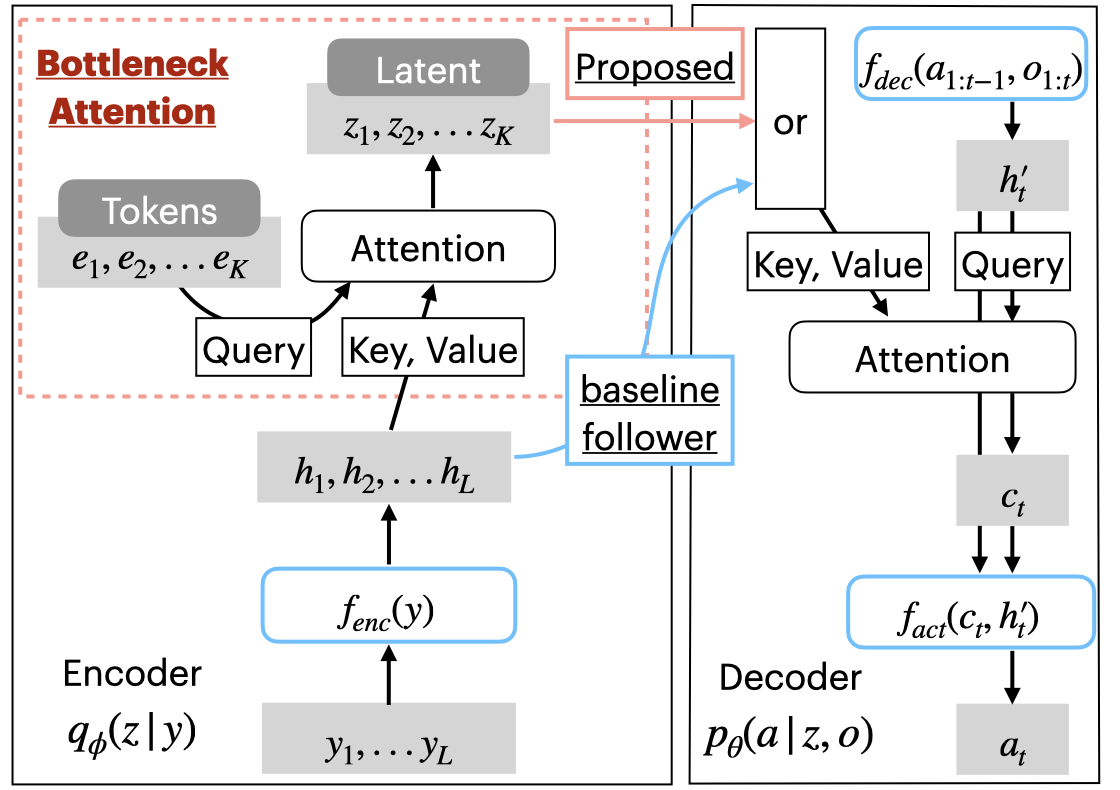}
  \caption{Schematic diagram of the proposed architecture.
  The baseline follower uses the hidden states $h$ in attention, whereas our architecture uses the latent variables $z$.
  }
  \label{image:architecture}
\end{figure}

Here we propose a novel module called bottleneck attention.
Also, we name the M-VAE with the module as MS-VAE.

While the attention mechanism plays an important role in seq2seq models, adapting it to M-VAE is not trivial because the data of different modalities mutually interact only through the shared representations, which are typically represented as a single vector.
For example, our follower (Eq. \ref{eq:rollout}) generates actions from $z$, not directly from a language instruction $y$ or its hidden state $h$.
Then, a naive approach to incorporate attention would be to sample latent variables for each timestep.
Specifically, this approach first samples the hidden states $[h_1, ..., h_L] = f_{enc} (y)$ as in the baseline follower, and then sample a latent variable $z_l$ from each $h_l$.
The resulting latent variables, $z = [z_1, ..., z_L]$, can be used as keys and values instead of $h$ by replacing the calculation of context by $c_t = f_{att} (h^\prime_t, z)$.
However, in this approach, the sequence lengths of the latent variables inferred from $q_\phi(z|a, o)$ and $q_\phi(z|y)$ become $T$ and $L$, respectively.
The discrepancy between the sequence lengths promotes a mismatch between $q_\phi(z|a, o)$ and $q_\phi(z|y)$, making the learning of shared representation difficult.
Moreover, some M-VAE objective functions \citep{suzuki2016joint,wu2018multimodal} are incompatible with the discrepancy.

As illustrated in the limitation of the naive approach, a critical challenge for the introduction of attention is to absorb the difference in sequence lengths of trajectory and language.
To avoid the discrepancy, we propose a novel module called bottleneck attention.
The module performs key-value attention to obtain $z = [z_1, ..., z_K]$, using $h$ as keys and values, and additional parameters $e = [e_1, ..., e_K]$ (referred to as tokens) as queries\footnote{
Strictly speaking, the module calculates the mean and variance of $q_\phi(z|a, o)$, which will be used in a reparameterization trick.}.
The tokens are model parameters and are initialized and optimized in a manner similar to processing word embeddings.
Here, $K$ denotes a sequential length, and each $z_i$ is a $D$-dimensional random variable.
$K$ must be specified as a hyperparameter and shared across the modalities, which enables the alignment of the sequence lengths.
Intuitively, this module extracts $K$ vectors that are important for the task from the variable-length hidden states $h = [h_1, ..., h_L]$.
The extracted latent variables are subsequently used in attention by replacing the calculation of context by $c_t = f_{att} (h^\prime_t, z)$, as in the naive approach.

An overview of the architecture is given in Figure \ref{image:architecture}.
As shown in the figure, the bottleneck attention can be attached to the baseline follower and is orthogonal to specific design for the modalities.
That is, the encoder $q_\phi(z|y)$ can be obtained by adding the bottleneck attention to the existing encoder of the follower $f_{enc}$.
In addition, for the decoder $p_\theta(a|z, o)$, the existing decoder of the follower, $f_{dec}$ and $f_{act}$, can be used without modification.
Conversely, although the follower is used as an example so far, it is also possible to introduce attention to the speaker using the same procedure.
Here we can use the existing encoder and decoder of the speaker as $q_\phi(z|a, o)$ and $p_\theta(y|z)$ of the MS-VAE.

\subsection{Domain Distance Regularization} \label{sec:prop:obj}

To further assist semi-supervised learning with MS-VAE, we propose a regularization term, which we call domain distance regularization.
Here, we denote the distribution of embeddings of the paired trajectories as $\rho = \mathbb{E}_{\tau \in D_{p}} [q_\phi(z | \tau)]$, and denote the distribution of the unpaired trajectories as $\upsilon = \mathbb{E}_{\tau^\prime \in D_{u}} [q_\phi(z | \tau^\prime)]$.

The regularization is motivated as follows.
Because $D_{p}$ and $D_{u}$ are used in different objective functions (Eqs. \ref{eq:elbo-mvaelf} and \ref{eq:traj-vae}), $\rho$ and $\upsilon$ are not necessarily the same.
In particular, because only action $a$ is decoded in Eq. \ref{eq:traj-vae}, the trajectory embeddings for $D_{u}$ could only acquire information about $a$ and ignore information about the observation $o$ and the language $y$ that relates to $o$.
Then, there will be a discrepancy between the distribution of the unpaired trajectory embeddings $\upsilon$ and the paired trajectory embeddings $\rho$.
This discrepancy indicates that DNNs trained using Eq. \ref{eq:traj-vae} overfit in trajectory reconstruction and do not generalize as a follower or speaker.

To alleviate this problem, we use a regularization term $D(\rho, \upsilon)$, which is the distance between $\rho$ and $\upsilon$.
For the specific choice of the distance, any existing estimation method, such as  adversarial learning or maximum mean discrepancy \citep{zhao2017infovae} can be used.
In our experiments, we used the sliced Wasserstein distance \citep{kolouri2018swae}, which was shown to have robustness (see Details in Appendix B).

Using the regularization term and Eq. \ref{eq:elbo-sum}, the final objective function of the proposed method is the weighted sum of the following terms:.
\begin{eqnarray}
    \max \mathbb{E}_{\{\tau, y\} \in D_{p}} [\bar{\mathcal{J}}] + \gamma \mathbb{E}_{\tau \in D_{u}} [\mathcal{V}] - \alpha D(\rho, \upsilon). \label{eq:elbo-final}
\end{eqnarray}
Here, $\gamma$ and $\alpha$ are the weighting parameters.

\section{Related Works} \label{sec:related}

The speaker-follower model is a dominant approach to utilize unpaired trajectories, and many subsequent studies \citep{tan2019envdrop,yu2020take,fu2020counterfactual,huang2019multi} have been proposed.
For example, \citet{yu2020take,fu2020counterfactual} proposed methods for the collection of unpaired trajectory data with higher quality.
\citet{huang2019multi} proposed using a discriminator to screen out incorrect instructions generated by a speaker.
In addition, in RL settings, \citet{cideron2020higher} proposed a data augmentation method that is very similar to the speaker-follower model.
While the speaker-follower model and the subsequent studies use only paired data to train the speaker model, the proposed method utilizes unpaired trajectory data.
Therefore, the proposed method is complementary to the speaker-follower model because it improves the speaker in a semi-supervised manner.

Our method is also closely related to \citet{artetxe2018unsupervised,lample2018unsupervised}.
These studies proposed two techniques to utilize unpaired data in the literature on machine translation: (i) back-translation and (ii) reconstruction via shared representation.
Among them, (ii) is very similar to $\mathcal{V}$ in Eq. \ref{eq:traj-vae} because both methods learn shared representation and perform reconstruction via it.
However, although the training procedures are similar, our method differs in how the shared representation is learned:
it learns the shared representation in the framework of multimodal generative models, assisted by the bottleneck attention module.
Also, note that the speaker-follower model is based on (i) back-translation.
Therefore, the complementarity between our method and the speaker-follower model is supported by the results obtained in the machine translation domain.

Another approach to utilizing auxiliary data for instruction following is to use pretrained transformers.
This approach was originally proposed in the language domain \citep{devlin2019bert} and applied to instruction following (e.g., \citet{hong2021vln}).
In addition, many studies have proposed methods that do not utilize auxiliary data, such as the use of data augmentation by adding noise \citep{tan2019envdrop} or the use of auxiliary loss (e.g., \citet{wang2019reinforced}).
By contrast, our study introduces the novel semi-supervised approach.

From the perspective of model architecture, a module similar to bottleneck attention was adopted by \citet{lee2019set} to extract a fixed number of important elements from the features.
While these studies used the module for feature extraction, our method differs in its use in the introduction of attention to cross-modality conversion with M-VAE.
In addition, to perform inference on latent variables with sequential structure as in this study, \citet{kovcisky2016semantic} proposed the use of a seq2seq model that outputs variable-length latent variables.
While their method requires an additional RNN to generate the latent variables, our proposed method uses only the additional attention, which enables parallel computation and is less likely to cause gradient vanishing or explosion.

\section{Experiments} \label{sec:exp}

\subsection{Settings}

Experiments were conducted in two environments: \textbf{BabyAI} \citep{chevalier-boisvert2018babyai} and \textbf{R2R} \citep{anderson2018vision}.
In BabyAI, four tasks were used: \textbf{GoToSeq}, \textbf{BossLevel}, \textbf{GoToSeqLocal}, and \textbf{BossLocal}.
GoToSeqLocal and BossLocal are the miniaturized versions of GoToSeq and BossLevel, respectively, and were created in our study.
In the GoToSeq and GoToSeqLocal tasks, one subgoal is to ``reach the location of a specific object'' (described as a GoTo instruction), and up to four subgoals are given as an instruction.
The BossLevel and BossLocal tasks are similar in that up to four subgoals are given as an instruction; however, they have various types of subgoals in addition to GoTo instruction, such as ``pick up a specific object''.
The details of these tasks including instruction samples are presented in Appendix C.
The goal of the follower is to accomplish these instructions, and the performance is evaluated by the mean \textbf{success rate (SR)} for randomly sampled test tasks.
In addition, the \textbf{BLEU-4} score, which is commonly used in studies on machine translation, was used to evaluate the speaker.

In R2R, the objective of the follower is to accomplish instructions that describe the path to the goal.
This dataset contains 7189 trajectories, and each trajectory has manually annotated language instructions.
In addition, for semi-supervised learning, we used the 178,000 unpaired trajectory data created by \citet{daniel2018speakerfollower}.
For evaluation, \textbf{SR}, \textbf{oracle success rate (OSR)}, and \textbf{navigation error (NE)} were used as in previous studies \citep{anderson2018vision,daniel2018speakerfollower}.
When evaluating SR, the follower must judge whether it has reached the goal.
By contrast, when evaluating OSR, the judgment is done automatically.
NE is the distance between the final location of the follower and the true goal position.
In addition, the R2R has three environments for validation: \textbf{validation seen}, \textbf{validation unseen}, and \textbf{test}, and the scores are measured in each of them.
Details of the experimental settings are described in Appendix D.

\subsection{Effect of Each Component of MS-VAE}

\paragraph{Comparing Architectures}

\begin{table}
  \caption{Success rate (\%) for different architectures.
  Bolded scores indicate the first or second highest within each task.
  }
  \label{tab:compare-arch}
  \begin{center}
    \scalebox{0.72}{
      \begin{tabular}{llrrrr}
      \toprule
                      &  &  GoToSeqLocal &  GoToSeq &  BossLocal &  BossLevel \\
      Architecture & $K$ &               &          &                 &            \\
      \midrule
      seq2seq &  &              99.3 &     \textbf{96.3}       &         89.8 &       49.5 \\
      w/ attention &  &        \textbf{99.8} &  95.0 & \textbf{99.8} &      \textbf{89.6} \\
      w/ bottleneck & 4 &          \textbf{99.6} &  94.0          &         97.2 &       82.3 \\
                     & 16 &           99.3 &      \textbf{96.2} &                  \textbf{99.2} &      \textbf{87.1} \\
      \bottomrule
      \end{tabular}
    }
  \end{center}
\end{table}

Here, we compare models with different architectures to confirm the contribution of the proposed bottleneck attention.
Specifically, we compare the seq2seq model without attention (seq2seq in Table \ref{tab:compare-arch}), the seq2seq model with attention (w/ attention), and the proposed model with bottleneck attention (w/ bottleneck).
In addition, we evaluated the influence of the hyperparameter $K$, the number of tokens, on the performance of the proposed model.
For a fair comparison, we trained all these models with the same IL loss, using a sufficient sample size of paired data ($M$ equals one million).

Table \ref{tab:compare-arch} shows that the presence of attention does not affect the success rate when the instructions are relatively simple as in GoToSeq.
By contrast, attention influences the success rate when the instructions are relatively complex, even if the environment is small, as in BossLocal.
These facts indicate that when complex language instructions were given, the action decoder could not process the whole of them at once, so giving attention may have contributed to the performance.
In addition, the proposed model achieves almost the same success rate as the model with attention.
Specifically, for BossLocal and BossLevel, the baseline seq2seq model reduces the success rates by approximately 10\% and 40\%, respectively, compared to the seq2seq model with attention, whereas the proposed model reduces them by only 0.6\% and 2.6\%.
Since the standard attention cannot be directly adopted in MS-VAE as described in Section \ref{sec:prop:attn}, these results suggest that the bottleneck attention, which rarely compromises the performance, is suitable as a component of MS-VAE.
In addition, because $K=16$ led to a higher success rate compared to $K=4$ in BossLocal and BossLevel, $z$ requires a somewhat large sequence length to embed complex instructions.
Here, note that we did not test $K$ that is larger than $16$ because the length of instruction is about $10$ as shown in Appendix C.

\paragraph{Contribution of Each Loss}

\begin{table}
  \caption{Success rates and BLEU scores in BabyAI for ablation study on the objective functions.
  }
  \label{tab:compare-parameter}
  \begin{center}
    \scalebox{0.76}{
      \begin{tabular}{lllrrrr}
      \toprule
            &     &       Task & \multicolumn{2}{l}{GoToSeqLocal} & \multicolumn{2}{l}{BossLocal} \\
            &     &       {} & SR & BLEU & SR & BLEU \\
      Method & $\mathcal{V}$ & $D(\rho, \upsilon)$ &                       &            &                       &            \\
      \midrule
      supervised &   &     &                                    55.0 &      10.43 &                 45.8 &       4.77 \\
      \hline
      MS-VAE &  &  &                              51.0 &      11.20 &                 42.0 &       6.47 \\
             & \checkmark  &   &                   68.3 &      10.86 &                 69.2 &       5.97 \\
      (full) & \checkmark  & \checkmark &     {\bf71.7} & {\bf11.73} &            {\bf76.7} &  {\bf7.42} \\
      \bottomrule
      \end{tabular}
    }
  \end{center}
\end{table}

\begin{table}
  \caption{Success rates for the baselines and the proposed two semi-supervised approaches.
  The values with ${}^*$ are adopted from the results reported in \citep{daniel2018speakerfollower}
  }
  \label{tab:boss-local}
  \scalebox{0.7}{
    \begin{tabular}{llllll}
    \toprule
    &        & Follower    & Speaker    &  \multicolumn{2}{c}{SR} \\
    &        &  w/ $D_u$   & w/ $D_u$   &  BabyAI     & R2R \\
    &        &             &            &  -BossLocal & -unseen \\
    & Method &             &            &             &\\
    \midrule
                   & follower &           &           &                  45.3      & 31.2${}^*$ \\
                   & speaker-follower     & \checkmark &           &     75.6      & 35.5${}^*$ \\
    \hline
                   & MS-VAE-follower       & \checkmark &           &     76.7      & 33.3 \\
                   & MS-VAE-speaker-follower & \checkmark & \checkmark &  {\bf82.6} & {\bf39.7} \\
    \bottomrule
    \end{tabular}
  }
\end{table}

\begin{table*}
  \caption{Comparing the proposed method and the existing semi-supervised learning methods in R2R.
  $\downarrow$ ($\uparrow$) denotes that the lower (upper) is better.
  Bolded values indicate the best results.}
  \label{tab:r2r}
  \begin{center}
    \scalebox{0.9}{
      \begin{tabular}{lllll|lll|lll}
      \toprule
           &  & \multicolumn{3}{c}{Validation Seen} & \multicolumn{3}{c}{Validation Unseen} & \multicolumn{3}{c}{Test} \\
           & {} &    NE$\downarrow$ &    SR$\uparrow$ &   OSR$\uparrow$ &    NE$\downarrow$ &    SR$\uparrow$  &   OSR$\uparrow$ &    NE$\downarrow$ &    SR$\uparrow$ &   OSR$\uparrow$ \\
      Decoding & Method &       &       &       &       &       &       &       &       &       \\
      \midrule
      greedy & speaker-follower \citep{daniel2018speakerfollower} &  3.36 &  66.4 &  73.8 &  6.62 &  35.5 &  45.0 &     - &     - &     - \\
           & \citet{tan2019envdrop}                             &  3.99 &  62.1 &     - &  {\bf5.22} &  {\bf52.2} &     - &     - &  {\bf51.5} &     - \\
           & \citet{huang2019multi}                             &  5.00 &  50.4 &     - &  5.90 &  39.1 &     - &     - &     - &     - \\
           & \citet{yu2020take}                                 &  5.03 &  53.0 &  61.6 &  6.29 &  38.9 &  46.7 &     - &     - &     - \\
           & \citet{fu2020counterfactual}                  &  {\bf3.30} &  {\bf68.2} &  {\bf74.9} &  6.10 &  38.8 &  46.7 &  {\bf5.90} &  37.6 &  {\bf46.4} \\
           \cline{2-11}
           & MS-VAE &                                               3.72 &  64.4 &  73.2 &  6.35 &  39.7 &  {\bf49.0} &  6.35 &  37.9 &  46.2 \\
      \hline
      \hline
      pragmatic & speaker-follower \citep{daniel2018speakerfollower} &  3.08 &  70.1 &  78.3 &  4.83 &  54.6 &  65.2 &  4.87 &  53.5 &  63.9 \\
           \cline{2-11}
           & MS-VAE &  {\bf2.70} &  {\bf74.3} &  {\bf80.3} &  {\bf4.44} &  {\bf56.9} &  {\bf66.4} &  {\bf4.52} &  {\bf56.1} &  {\bf64.2} \\
      \bottomrule
      \end{tabular}
    }
  \end{center}
\end{table*}

Here, the models were trained on the GoToSeqLocal and BossLocal tasks in a semi-supervised setting, using a small amount of paired data ($M=1000$) and unpaired trajectory data ($N=$ one million).
Table \ref{tab:compare-parameter} shows the contributions of the two loss functions of the MS-VAE.
The first loss is $\mathcal{V}$ in Eq. \ref{eq:elbo-final}, i.e., whether to use unpaired trajectory data or not.
The second is $D(\rho, \upsilon)$ in Eq. \ref{eq:elbo-final},  i.e., whether to use the domain distance loss.
The hyperparameter settings for these losses can be found in Appendix D.
All methods were trained with three random seed trials, and the average scores were reported.
Here, because MS-VAE can be used as both follower and speaker, we used the SR and BLEU scores to evaluate the follower and speaker, respectively.
As the baseline method, we used the seq2seq model with attention trained on only paired data (``supervised'' in Table \ref{tab:compare-parameter}).

First, comparing MS-VAE with all loss functions (full) with supervised baseline and MS-VAE without $\mathcal{V}$ in Table \ref{tab:compare-parameter}, the full model has higher SR and BLEU.
This indicates that training with unpaired trajectories improves the SR and BLEU.
Note that the BLEU scores reported in the table are lower than typical scores in machine translation (e.g., $>20$) owing to the insufficient paired samples and the fact that the generation of languages from trajectories is an ill-posed problem.
However, the table shows a consistent, albeit modest, improvement in the BLEU scores (see Appendix D for statistical tests).
In addition, to evaluate the speaker, we validated the performance of the speaker-follower model in the next paragraph.
Second, assuming the use of $\mathcal{V}$, the regularization term ($D(\rho, \upsilon)$) improves BLEU scores.
This may be because the use of unpaired data promotes a mismatch between $\rho$ and $\upsilon$, which are the embedding distributions of paired and unpaired trajectories created by the speaker encoder $q_\phi (z|a, o)$, and degrades the generalization performance of the speaker;
but the regularization may mitigate the speaker degradation caused by the domain mismatch.

\paragraph{Comparing Two Semi-Supervised Approaches} \label{sec:exp:two-approach}

In Table \ref{tab:boss-local}, we evaluated the two approaches described in Section \ref{sec:prop:pgm}: MS-VAE-follower and MS-VAE-speaker-follower, using BabyAI's BossLocal task and R2R's validation unseen.
In addition, as baseline methods, we used the seq2seq model with attention (follower) and the speaker-follower model.
To clarify the relationship between these methods, Table \ref{tab:boss-local} shows whether the unpaired trajectory data are used to train the follower (follower w/ $D_u$) and whether the data are used to train the speaker (speaker w/ $D_u$).
As shown in the table, the performance of the MS-VAE-follower outperformed that of the follower.
This shows that the MS-VAE follower can be used as a novel semi-supervised learning approach.
Second, the MS-VAE speaker-follower achieved the best performance among the various methods.
This indicates that the speaker-follower model can be improved using the trained MS-VAE as a speaker, i.e., the proposed method is complementary to the speaker-follower model.

\subsection{Comparison with Previous Methods}

Here, we compare MS-VAE with previous methods for semi-supervised instruction following, using the R2R environment.
MS-VAE-speaker-follower was used as our proposed method.
The method was compared with the speaker-follower \citep{daniel2018speakerfollower} and subsequent methods \citep{tan2019envdrop,huang2019multi,yu2020take,fu2020counterfactual}.
In addition, because the speaker model can be used not only for data augmentation but also for a beam-search method called pragmatic inference \citep{daniel2018speakerfollower}, we reported the scores for the proposed method with the pragmatic inference\footnote{Although the other methods \citep{tan2019envdrop,huang2019multi,yu2020take,fu2020counterfactual} could adopt pragmatic inference in principle, the scores for pragmatic inference were not reported in most cases, thus they are not listed in Table \ref{tab:r2r}.}.
As shown in Table \ref{tab:r2r}, MS-VAE outperforms the speaker-follower model in the validation unseen and test environments.
Moreover, with pragmatic inference, the performance of MS-VAE is superior to that of the speaker-follower in all environments, probably because the speaker plays a more significant role when using pragmatic inference.
In addition, the proposed method achieves almost the same level of performance as the other methods.
One exception is that \citet{tan2019envdrop} achieved significantly high SR in the validation unseen and test environments, which can be attributed to their use of reinforcement learning.
Finally, note that the proposed method can be complementary to these baseline methods since the proposed method improves the performance of the speaker by using unpaired trajectory data as described in Section \ref{sec:related}.

\section{Conclusion}

In this study, we proposed MS-VAE for a generative model-based semi-supervised learning in sequence-to-sequence tasks.
Compared to the existing M-VAE, MS-VAE can employ an attention mechanism, and its importance in the instruction-following task is confirmed using the results in Table \ref{tab:compare-arch}.
In addition, Table \ref{tab:compare-parameter} shows that MS-VAE, along with domain distance regularization, can improve the speaker.
In addition, Tables \ref{tab:compare-parameter} and \ref{tab:boss-local} show that the proposed model can improve the follower by using unpaired trajectories.
Moreover, Tables \ref{tab:boss-local} and \ref{tab:r2r} show that the MS-VAE-based speaker improved the performance of the speaker-follower model.
These results indicate that the generative model-based approach contributes to instruction following in two ways, and is complementary to the speaker-follower model.

The proposed model may easily incorporate unpaired language data into training by replacing Eq. \ref{eq:traj-vae} with the language model $p_\theta(z, y)$, but it remains future work.
Other future work may apply MS-VAE to other applications that target multimodal sequential data such as image-captioning and text-to-speech because the proposed module is orthogonal to specific design for the modalities.

\appendix

\section{Lower Bound on $\mathcal{J}$}

Because $\mathcal{J}$ (the objective function for paired data) is intractable for analytical computations, we use the following lower bound for optimization.
\begin{eqnarray}
    \mathcal{J} \geq \bar{\mathcal{J}} =& \frac{1}{2} \sum_{m=1}^2 [ \mathcal{A}_m + \mathcal{B}_m + \mathcal{C}_m ], \label{eq:elbo-mvaelf-prop} \\
                \mathrm{where} \;\; \mathcal{A}_m =& \mathbb{E}_{q_\phi(z|x_{(m)})} [\log p_\theta(x_{(m)}|z)], \nonumber \\
                                    \mathcal{B}_m =& - D_{\mathrm{KL}} [q_\phi(z|x_{(m)})|p_\theta(z)], \nonumber \\
                                    \mathcal{C}_m =& \mathbb{E}_{q_\phi(z|{(x_{(m)})})} [\log p_\theta(x_{(n \neq m)}|z)]. \nonumber
\end{eqnarray}
In Eq. \ref{eq:elbo-mvaelf-prop}, the notation $x_{(1)} = a | o$ and $x_{(2)} = y$ were used for better visibility.
$\mathcal{A}_m$ represents the negative reconstruction error for modality $m$.
$\mathcal{B}_m$ is the Kullback-Leibler (KL) distance between the variational posterior and prior.
Note that the KL term makes the trajectory embedding $q_\phi(z|x_{(1)} )$ and the language embedding $q_\phi(z|x_{(2)} )$ closer to the same prior distribution, which may facilitate the learning of the shared representation.
$\mathcal{C}_m$ is the negative error for cross-modal conversion, which can be computed by using the generative model as follower and speaker.

The derivation of Eq. \ref{eq:elbo-mvaelf-prop} is given as follows:
\begin{eqnarray}
    \mathcal{J} =& \frac{1}{2} \sum_{m=1}^2 \mathbb{E}_{q_\phi(z|x_{(m)})} [\log \frac{p(z, x_{(1)}, x_{(2)})}{q_\phi(z|x_{(1)}, x_{(2)})} ] \nonumber \\
                =& \frac{1}{2} \sum_{m=1}^2 \mathbb{E}_{q_\phi(z|x_{(m)})} [\log p_\theta(x_{(1)}|z) + \log p_\theta(x_{(2)}|z) \nonumber  \\
                 &+ \log p_\theta(z) - \log q_\phi(z|x_{(1)}, x_{(2)})] \nonumber \\
                =& \frac{1}{2} \sum_{m=1}^2 \mathbb{E}_{q_\phi(z|x_{(m)})} [\log p_\theta(x_{(m)}|z)] \nonumber  \\
                 &+ \frac{1}{2} \sum_{m=1}^2 \mathbb{E}_{q_\phi(z|x_{(m)})} [\log p_\theta(x_{n \neq m}|z)] \nonumber  \\
                 &+ \frac{1}{2} \sum_{m=1}^2 \mathbb{E}_{q_\phi(z|x_{(m)})}[\log \frac{p_\theta(z)}{q_\phi(z|x_{(1)}, x_{(2)})} \frac{q_\phi(z|x_{(m)})}{q_\phi(z|x_{(m)})} ] \nonumber \\
                =& \frac{1}{2} \sum_{m=1}^2 \mathbb{E}_{q_\phi(z|x_{(m)})} [\log p_\theta(x_{(m)}|z)] \nonumber  \\
                 &+ \frac{1}{2} \sum_{m=1}^2 \mathbb{E}_{q_\phi(z|x_{(m)})} [\log p_\theta(x_{n \neq m}|z)] \nonumber  \\
                 &+ \frac{1}{2} \sum_{m=1}^2 D_{\mathrm{KL}} [q_\phi(z | x_{(m)})|q_\phi(z | x_{(1)}, x_{(2)})]  \nonumber \\
                 &- \frac{1}{2} \sum_{m=1}^2 D_{\mathrm{KL}} [q_\phi(z | x_{(m)})|p_\theta(z)] \nonumber \\
             \geq& \frac{1}{2} \sum_{m=1}^2 \mathbb{E}_{q_\phi(z|x_{(m)})} [\log p_\theta(x_{(m)}|z)] \nonumber  \\
                 &+ \frac{1}{2} \sum_{m=1}^2 \mathbb{E}_{q_\phi(z|x_{(m)})} [\log p_\theta(x_{n \neq m}|z)] \nonumber  \\
                 &- \frac{1}{2} \sum_{m=1}^2 D_{\mathrm{KL}} [q_\phi(z | x_{(m)})|p_\theta(z)] \nonumber \\
                =& \frac{1}{2} \sum_{m=1}^2 [\mathcal{A}_m + \mathcal{B}_m + \mathcal{C}_m] = \bar{\mathcal{J}}.
\end{eqnarray}

Also, Eq. \ref{eq:elbo-mvaelf-prop} might also be computed using a sampling method such as the DReG Estimator \cite{tucker2018doubly} as was done in \cite{shi2019variational}.
Although the comparison of the sampling methods and the variational lower bound (Eq. \ref{eq:elbo-mvaelf-prop}) is beyond the scope of this study, the variational lower bound is advantageous in that it is more efficient in graphics processing unit (GPU) memory because it does not require the sampling of potentially high-dimensional trajectories.

\begin{figure}[t]
  \centering
  \includegraphics[width=\columnwidth]{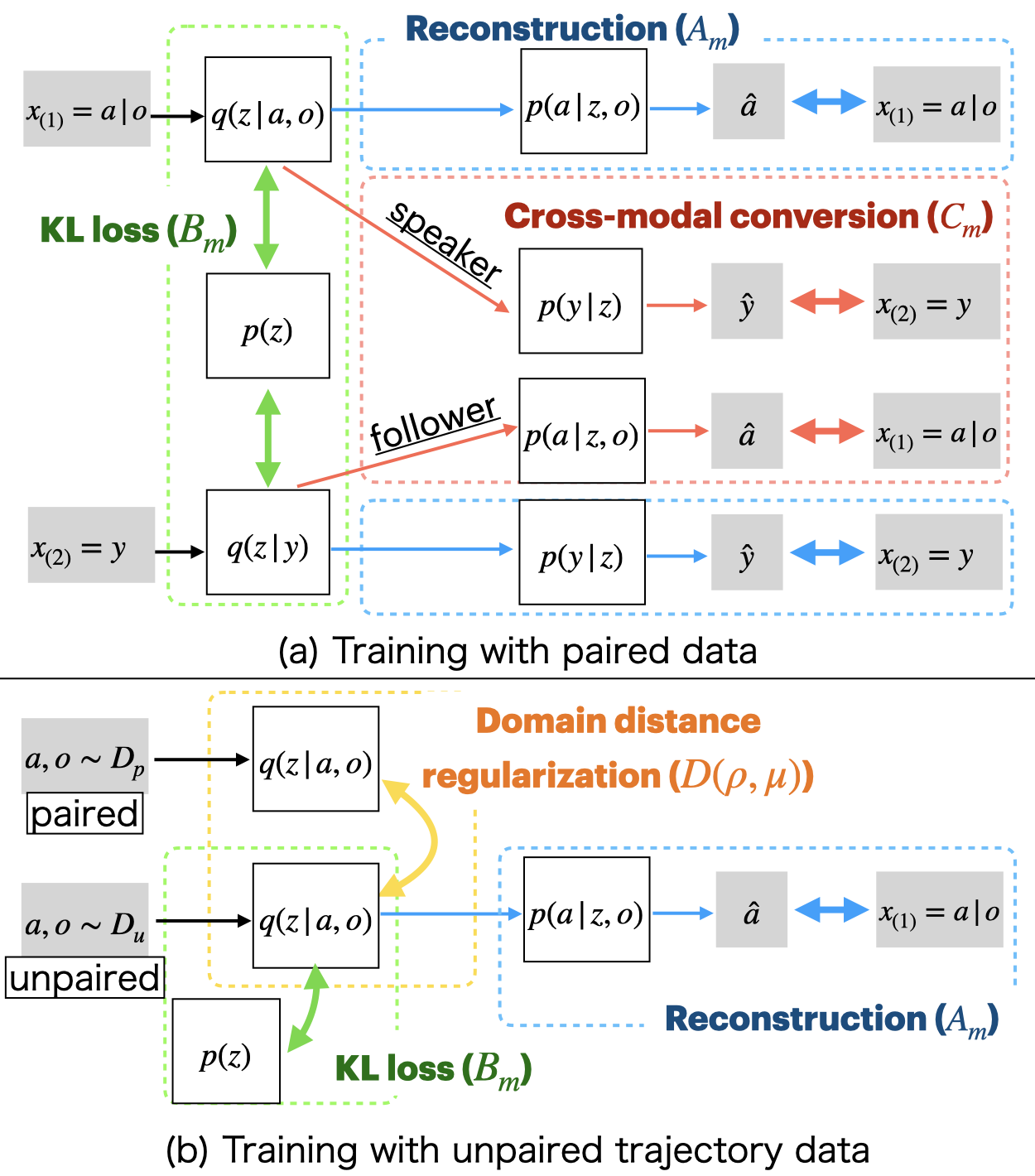}
  \caption{Schematic diagram for objective functions of (a) training with paired data and (b) training with unpaired trajectories including domain distance regularization.
  $\hat{a}$ and $\hat{y}$ denote the samples generated from the model.
  }
  \label{image:loss}
\end{figure}

Note that, using the notation $\mathcal{A}_m$ and $\mathcal{B}_m$, $\mathcal{V}$ (the objective function for unpaired data) can be represented as follows:
\begin{eqnarray}
    \mathcal{V}  =&& \mathbb{E}_{q_\phi(z|a, o)} [\log p_\theta(a|z, o)] + D_{\mathrm{KL}} [q_\phi(z|a, o)|p_\theta(z)] \nonumber \\
                 =&& \mathcal{A}_1 + \mathcal{B}_1.
\end{eqnarray}
Therefore, the objective functions for paired and unpaired data can be illustrated as in Figure \ref{image:loss}.

\section{Implementation Details on Domain Distance Regularization} \label{app:ddr}

In our experiments, we adopted the sliced Wasserstein distance \cite{kolouri2018swae} for the domain distance regularization $D(\rho, \upsilon)$, and also used two additional techniques to stabilize the training.
First, instead of measuring the distance for the entire $z = [z_1, ..., z_K]$ of the dimension $K \times D$, we calculated the distance for each $z_k$ and used their sum, which reduces the dimension for each calculation.
Second, instead of performing random sampling from $q_\phi(z | a, o)$, we use the mean values $\mu_z=[\mu_1, ..., \mu_K]$ of the distribution $q_\phi(z | a, o)$ obtained using the reparameterization trick because $\mu_z$ is used as input to the speaker and follower (see Section 3.1).
That is, the following $D^\prime$ was used as a regularization term in the experiment:
\begin{eqnarray}
    D^\prime &= \sum_{k=1}^K D(\rho_k, \upsilon_k), \nonumber \\
    \mathrm{where}& \;\; \rho_k = \mathbb{E}_{a,o \in D_{p}} [\mu_k], \upsilon_k = \mathbb{E}_{a,o \in D_{u}} [\mu_k] \nonumber
\end{eqnarray}
Moreover, the sliced wasserstein distance has a hyperparameter, the number of random projections (see Algorithm 1 in \citet{kolouri2018swae}), which was set to $50$ as used in the official implementation\footnote{https://github.com/skolouri/swae} of \citet{kolouri2018swae}.

\section{Details on BabyAI Tasks} \label{app:babyai_tasks}

\begin{figure*}[t]
  \begin{center}
      \begin{tabular}{cc}
      \begin{minipage}[l]{0.5\hsize}
        \subfigure[GoToSeq]{\includegraphics[width=\columnwidth]{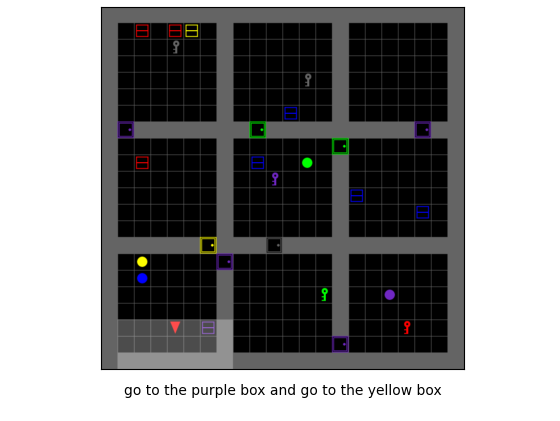}
        }
      \end{minipage}
      \begin{minipage}[l]{0.5\hsize}
        \subfigure[BossLevel]{\includegraphics[width=\columnwidth]{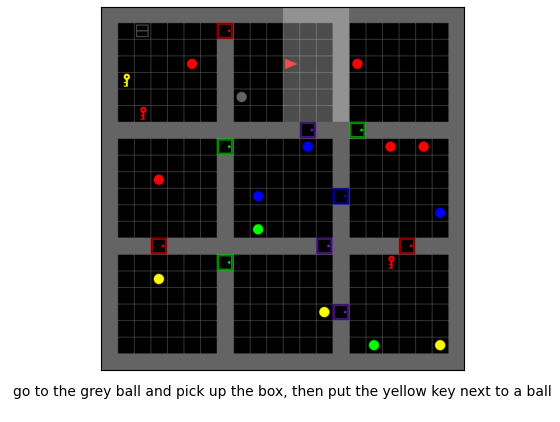}
        }
      \end{minipage}
      \\
      \begin{minipage}[l]{0.5\hsize}
        \subfigure[GoToSeqLocal]{\includegraphics[width=\columnwidth]{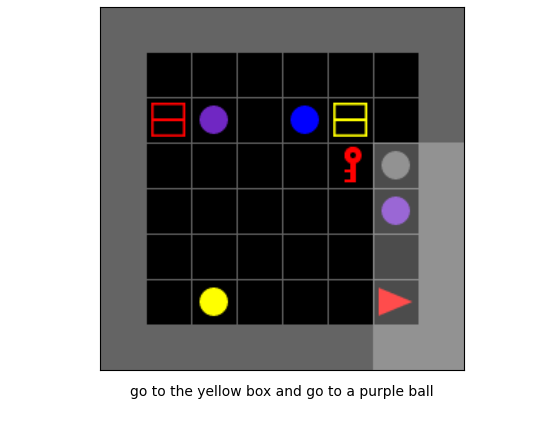}
        }
      \end{minipage}
      \begin{minipage}[l]{0.5\hsize}
        \subfigure[BossLocal]{\includegraphics[width=\columnwidth]{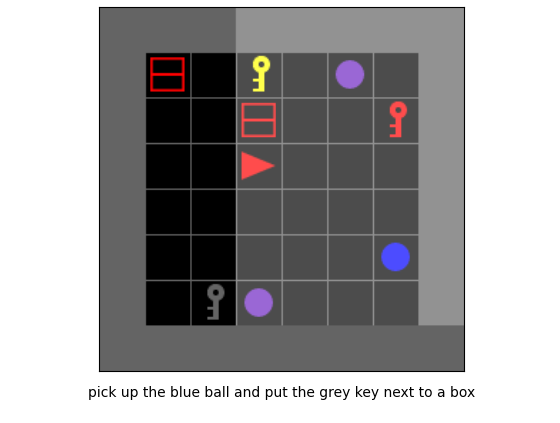}
        }
      \end{minipage}
    \end{tabular}
    \caption{Examples of four Baby AI tasks.
    Instructions are at the bottom of each image
    }
    \label{fig:babyai-tasks}
   \end{center}
\end{figure*}

Figure \ref{fig:babyai-tasks} shows examples of four BabyAI tasks used in our experiments: GoToSeq, BossLevel, GoToSeqLocal, and BossLocal.
Among them, GoToSeqLocal and BossLocal correspond to miniaturized versions of the GoToSeq and BossLevel tasks, respectively, and were created in our study.
Specifically, the gridworld of GoToSeq and BossLevel consists of nine rooms (3 $\times$ 3), whereas the gridworld of GoToSeqLocal and BossLocal consists of only one room.
GoToSeq and GoToSeqLocal, and BossLevel and BossLocal have almost the same language instructions, except that the instructions of ``opening the door'' do not appear in the miniaturized tasks because there are no doors in the tasks.
In addition, Table \ref{tab:babyai-stats} shows the average length of trajectories and language instructions for each task in BabyAI.
As shown in the table, the sequence length of the trajectories is much longer in GoToSeq and BossLevel, which is attributed to their larger environments.
Also, GoToSeqLocal and BossLocal (or GoToSeq and BossLevel) have the almost same instruction and trajectory lengths.
In addition, they have no difference in the environment (e.g., they have the same number of objects in a room).
Therefore, we believe that the only striking difference between these two settings is the complexity of tasks.
The implementations of GoToSeqLocal and BossLocal are available in our code submitted as supplementary materials.

\begin{table}
  \caption{Average length of trajectories and language instructions for each task of BabyAI.
  }
  \label{tab:babyai-stats}
  \begin{center}
    \scalebox{0.7}{
      \begin{tabular}{lrrrr}
        \toprule
        Task &  GoToSeqLocal &  GoToSeq &  BossLocal &  BossLevel \\
        \midrule
        Instruction Length &         10.56 &    10.78 &          11.57 &      12.41 \\
        Trajectory Length  &          8.85 &    73.06 &          12.07 &      84.09 \\
        \bottomrule
      \end{tabular}
    }
  \end{center}
\end{table}

\section{Details on the Experiments} \label{app:exp_details}

\subsection{Architecture and Training Settings}

In the BabyAI experiments, the neural network architecture of the baseline follower model (``w/ attention'' in Table 1 and ``supervised'' in Table 2) is exactly the same as that used in \citet{hui2020babyai}.
However, in the miniaturized GoToSeqLocal and BossLocal tasks, the dimension of the hidden state of the decoder LSTM is halved.
Conversely, the baseline speaker model (``supervised'' in Table 2) encodes trajectories (observations and actions) based on the gated recurrent unit (GRU).
Specifically, the observations $o_t$ were first mapped to feature vectors by convolutional layers that were used in \citet{hui2020babyai}, and the actions $a_t$ were also mapped to vectors using a linear transformation.
Subsequently, these feature vectors were used as input to the GRU.
The language was then generated using a two-layer transformer decoder.
For the architecture of the proposed method, as described in Section 3.2, the bottleneck attention module was added to the above follower and speaker.
The dimension of the latent variable was set to 128, which is the same as that of the language embedding in the implementation of \citet{hui2020babyai}.
The number of tokens was set to $K=4$ unless otherwise mentioned.

 Adam optimizer was used in the optimization as in \citet{hui2020babyai}.
For GoToSeq and BossLevel, the models were trained for 200 epochs with a batch size of 128, where each epoch consisted of 400 parameter updates (iterations).
For GoToSeqLocal and BossLocal, the models were trained for 200 epochs with a batch size of 256, where each epoch consisted of 200 parameter updates (iterations).
To speed up the convergence of MS-VAE, part of its parameters (language encoder and action decoder) were initialized with the trained baseline follower model.
Specifically, we pretrained $f_{enc}$, $f_{dec}$, and $f_{act}$ shown in Figure 1, and fine-tuned them by adding bottleneck attention.
In addition, the speaker in the speaker-follower model in BabyAI is used only in data augmentation and not in pragmatic inference.
As regards the measurement of SR, it is not suitable to use a hold-out validation, which is commonly used in supervised learning because the evaluation of SR requires a trial in the environment.
Subsequently, we measured the SR for each epoch, applied smoothing with a window size of five to eliminate the effect of outliers, and reported the largest value.
For the measurement of the BLEU score, we prepared validation and test data separately from the training data and applied the hold-out method.

In the R2R experiments, the architectures of the baseline speaker and follower models were the same as those in \citet{daniel2018speakerfollower}.
The architecture of the proposed method is the same as those of the baseline speaker and follower models, except for the additional bottleneck attention module.
Here, the number of tokens was set to $K=10$, and the dimension of latent variables was set to $D=512$, which is the same as the hidden state of the LSTM for the follower and speaker.

We trained the proposed method for 50000 and 20000 iterations when used as a follower and a speaker, respectively.
These number of iterations are the same as those used in \citet{daniel2018speakerfollower}.
Other hyperparameters related to the optimization, such as the learning rate, were the same as in the speaker and follower models used in \citet{daniel2018speakerfollower}.
The parameters of the MS-VAE were initialized using the trained baseline follower and speaker models.
During the training of the follower, early stopping was performed based on the success rate for the validation-unseen environment, as reported in \citet{tan2019envdrop}.

As for the prior distribution $p_\theta(z= [z_1, . , z_K])$, it can be defined using the standard normal distribution as in the standard VAE, or it can be defined using an autoregressive prior as in \citet{hafner2019learning} to improve the expressiveness.
On the experiments, we defined is using an autoregressive prior as in \citet{hafner2019learning}, where GRUs with hidden size 128 (BabyAI) and 512 (R2R) were employed.

\subsection{Hyperparameter Settings} \label{app:exp_details:hyperparameter}

Eq. 10 shows that the objective function of the proposed method has two weighting parameters $\alpha$ and $\gamma$.
Additionally, we introduced the hyperparameter $\beta$ for $\bar{\mathcal{J}}$ and $\mathcal{V}$ to control the effect of the KL loss as follows:
\begin{eqnarray}
    \bar{\mathcal{J}} =& \frac{1}{2} \sum_{m=1}^2 [ \mathcal{A}_m + \beta \mathcal{B}_m + \mathcal{C}_m ], \nonumber \\
    \mathcal{V} =& \mathcal{A}_m + \beta \mathcal{B}_m. \nonumber
\end{eqnarray}
This technique is called $\beta$-VAE \cite{higgins2017beta}, which is often employed in VAEs with an autoregressive decoder to prevent the problem of posterior collapse, where a latent variable becomes uninformative \cite{alemi2018fixing}.

In our experiments, these values were heuristically set to $\alpha=1/20$, $\gamma=1e2$, and $\beta=1e-1$ for BabyAI, and $\alpha=1/2$, $\gamma=1e3$, and $\beta=1e-4$ for R2R.
The reason for setting $\beta < 1$ is that when $\beta$ was set to a large value (e.g., $\beta=1$) in our preliminary experiments, the posterior collapse resulted in performance degradation of the follower.
In our experiments, we used the same value of $\beta$ regardless of the modality $m$; however, setting the value of $\beta$ for each modality may contribute to improved performance.
In R2R, $\alpha=1/2$ was set to align the scales of $\mathcal{A}_m + \mathcal{B}_m$ in $\bar{\mathcal{J}}$ and $\mathcal{V}$; however, other values have not been fully tested.
In BabyAI, we employed $\alpha=1/20$ because we compared $\alpha=1/2$ and $\alpha=1/20$ in an informal preliminary experiment and found that $\alpha=1/20$ achieved a slightly better BLEU score.
For $\gamma$, our preliminary experiments with one random seed trial showed a minor change in speaker and follower performance when using small values (e.g., $\gamma <= 1$).
By contrast, large heuristically chosen values ($\gamma=1e2$ and $\gamma=1e3$) showed performance improvements.

\subsection{Computing Infrastructures and Softwares}

For the GPUs, we used Quadro RTX8000, Tesla V100, or GeForce GTX 1080 Ti, depending on the availability of our server.
The GPUs all have 48 GB or 16 GB of memory.
In addition, all experiments were performed on a single GPU.
For the OS, CentOS Linux 7.5 or Ubuntu 16.04 was used for BabyAI, depending on the availability of the server.
For R2R, Ubuntu 18.04 was used.
The implementation was based on PyTorch \cite{paszke2019pytorch} version 1.4.0.

\subsection{Statistical Tests} \label{app:exp_details:test}

Here, we describe the statistical tests performed to confirm whether there are significant differences between the results of the experiments in Section 5.
First, in the experiments in Table 2 and the BossLocal task in Table 3, all methods were trained using three random seeds.
By contrast, the other experiments were run on a single random seed trial because it takes approximately a week to train a single model in a large-scale environment such as BabyAI's BossLevel or R2R.
Therefore, the tests were performed based on the results in Table 2.

Statistical tests were performed to compare ``supervised learning'' with ``MS-VAE (full),'' and ``MS-VAE trained on paired data only'' with ``MS-VAE (full).''
One-sided t-tests were performed independently for each environment (GoToSeqLocal and BossLocal).
The null hypothesis of this test is that ``there is no difference in the means of the scores (for the three random seed trials) between the methods.''
For the success rate and BLEU score, we obtained p-values less than 0.05 for all the statistical tests.
These results indicate that the proposed method improves the performance of followers and speakers by using unpaired trajectory data.

\clearpage
\bibliographystyle{named}
\bibliography{../ref}

\begin{thebibliography}{}

\bibitem[\protect\citeauthoryear{Alemi \bgroup \em et al.\egroup
  }{2018}]{alemi2018fixing}
Alexander Alemi, Ben Poole, Ian Fischer, Joshua Dillon, Rif~A Saurous, and
  Kevin Murphy.
\newblock Fixing a broken elbo.
\newblock In {\em International Conference on Machine Learning}, pages
  159--168, 2018.

\bibitem[\protect\citeauthoryear{Anderson \bgroup \em et al.\egroup
  }{2018}]{anderson2018vision}
Peter Anderson, Qi~Wu, Damien Teney, Jake Bruce, Mark Johnson, Niko
  S{\"u}nderhauf, Ian Reid, Stephen Gould, and Anton Van Den~Hengel.
\newblock Vision-and-language navigation: Interpreting visually-grounded
  navigation instructions in real environments.
\newblock In {\em Proceedings of the IEEE Conference on Computer Vision and
  Pattern Recognition}, pages 3674--3683, 2018.

\bibitem[\protect\citeauthoryear{Artetxe \bgroup \em et al.\egroup
  }{2018}]{artetxe2018unsupervised}
Mikel Artetxe, Gorka Labaka, Eneko Agirre, and Kyunghyun Cho.
\newblock Unsupervised neural machine translation.
\newblock In {\em International Conference on Learning Representations}, 2018.

\bibitem[\protect\citeauthoryear{Beetz \bgroup \em et al.\egroup
  }{2011}]{michael2011pancake}
Michael Beetz, Ulrich Klank, Ingo Kresse, Alexis Maldonado, Lorenz
  Mösenlechner, Dejan Pangercic, Thomas Rühr, and Moritz Tenorth.
\newblock Robotic roommates making pancakes.
\newblock In {\em 2011 11th IEEE-RAS International Conference on Humanoid
  Robots}, pages 529--536, 2011.

\bibitem[\protect\citeauthoryear{Chevalier-Boisvert \bgroup \em et al.\egroup
  }{2019}]{chevalier-boisvert2018babyai}
Maxime Chevalier-Boisvert, Dzmitry Bahdanau, Salem Lahlou, Lucas Willems,
  Chitwan Saharia, Thien~Huu Nguyen, and Yoshua Bengio.
\newblock Baby{AI}: First steps towards grounded language learning with a human
  in the loop.
\newblock In {\em International Conference on Learning Representations}, 2019.

\bibitem[\protect\citeauthoryear{Cideron \bgroup \em et al.\egroup
  }{2020}]{cideron2020higher}
Geoffrey Cideron, Mathieu Seurin, Florian Strub, and Olivier Pietquin.
\newblock Higher: Improving instruction following with hindsight generation for
  experience replay.
\newblock In {\em 2020 IEEE Symposium Series on Computational Intelligence
  (SSCI)}, pages 225--232. IEEE, 2020.

\bibitem[\protect\citeauthoryear{Devlin \bgroup \em et al.\egroup
  }{2019}]{devlin2019bert}
Jacob Devlin, Ming-Wei Chang, Kenton Lee, and Kristina Toutanova.
\newblock Bert: Pre-training of deep bidirectional transformers for language
  understanding.
\newblock In {\em Proceedings of the 2019 Conference of the North American
  Chapter of the Association for Computational Linguistics: Human Language
  Technologies, Volume 1 (Long and Short Papers)}, pages 4171--4186, 2019.

\bibitem[\protect\citeauthoryear{Fried \bgroup \em et al.\egroup
  }{2018}]{daniel2018speakerfollower}
Daniel Fried, Ronghang Hu, Volkan Cirik, Anna Rohrbach, Jacob Andreas,
  Louis-Philippe Morency, Taylor Berg-Kirkpatrick, Kate Saenko, Dan Klein, and
  Trevor Darrell.
\newblock Speaker-follower models for vision-and-language navigation.
\newblock In {\em Advances in Neural Information Processing Systems},
  volume~31. Curran Associates, Inc., 2018.

\bibitem[\protect\citeauthoryear{Fu \bgroup \em et al.\egroup
  }{2020}]{fu2020counterfactual}
Tsu-Jui Fu, Xin~Eric Wang, Matthew~F Peterson, Scott~T Grafton, Miguel~P
  Eckstein, and William~Yang Wang.
\newblock Counterfactual vision-and-language navigation via adversarial path
  sampler.
\newblock In {\em European Conference on Computer Vision}, pages 71--86.
  Springer, 2020.

\bibitem[\protect\citeauthoryear{Hafner \bgroup \em et al.\egroup
  }{2019}]{hafner2019learning}
Danijar Hafner, Timothy Lillicrap, Ian Fischer, Ruben Villegas, David Ha,
  Honglak Lee, and James Davidson.
\newblock Learning latent dynamics for planning from pixels.
\newblock In {\em International Conference on Machine Learning}, pages
  2555--2565. PMLR, 2019.

\bibitem[\protect\citeauthoryear{Higgins \bgroup \em et al.\egroup
  }{2017}]{higgins2017beta}
Irina Higgins, Loic Matthey, Arka Pal, Christopher Burgess, Xavier Glorot,
  Matthew Botvinick, Shakir Mohamed, and Alexander Lerchner.
\newblock beta-vae: Learning basic visual concepts with a constrained
  variational framework.
\newblock In {\em International Conference on Learning Representations}, 2017.

\bibitem[\protect\citeauthoryear{Hong \bgroup \em et al.\egroup
  }{2021}]{hong2021vln}
Yicong Hong, Qi~Wu, Yuankai Qi, Cristian Rodriguez-Opazo, and Stephen Gould.
\newblock Vln bert: A recurrent vision-and-language bert for navigation.
\newblock In {\em Proceedings of the IEEE/CVF Conference on Computer Vision and
  Pattern Recognition}, pages 1643--1653, 2021.

\bibitem[\protect\citeauthoryear{Huang \bgroup \em et al.\egroup
  }{2019}]{huang2019multi}
Haoshuo Huang, Vihan Jain, Harsh Mehta, Jason Baldridge, and Eugene Ie.
\newblock Multi-modal discriminative model for vision-and-language navigation.
\newblock In {\em Proceedings of the Combined Workshop on Spatial Language
  Understanding (SpLU) and Grounded Communication for Robotics (RoboNLP)},
  pages 40--49, 2019.

\bibitem[\protect\citeauthoryear{Hui \bgroup \em et al.\egroup
  }{2020}]{hui2020babyai}
David Yu-Tung Hui, Maxime Chevalier-Boisvert, Dzmitry Bahdanau, and Yoshua
  Bengio.
\newblock Babyai 1.1.
\newblock {\em arXiv preprint arXiv:2007.12770}, 2020.

\bibitem[\protect\citeauthoryear{Ko{\v{c}}isk{\`y} \bgroup \em et al.\egroup
  }{2016}]{kovcisky2016semantic}
Tom{\'a}{\v{s}} Ko{\v{c}}isk{\`y}, G{\'a}bor Melis, Edward Grefenstette, Chris
  Dyer, Wang Ling, Phil Blunsom, and Karl~Moritz Hermann.
\newblock Semantic parsing with semi-supervised sequential autoencoders.
\newblock In {\em Proceedings of the 2016 Conference on Empirical Methods in
  Natural Language Processing}, pages 1078--1087, 2016.

\bibitem[\protect\citeauthoryear{Kolouri \bgroup \em et al.\egroup
  }{2019}]{kolouri2018swae}
Soheil Kolouri, Phillip~E. Pope, Charles~E. Martin, and Gustavo~K. Rohde.
\newblock Sliced wasserstein auto-encoders.
\newblock In {\em International Conference on Learning Representations}, 2019.

\bibitem[\protect\citeauthoryear{Lample \bgroup \em et al.\egroup
  }{2018}]{lample2018unsupervised}
Guillaume Lample, Alexis Conneau, Ludovic Denoyer, and Marc'Aurelio Ranzato.
\newblock Unsupervised machine translation using monolingual corpora only.
\newblock In {\em International Conference on Learning Representations}, 2018.

\bibitem[\protect\citeauthoryear{Lee \bgroup \em et al.\egroup
  }{2019}]{lee2019set}
Juho Lee, Yoonho Lee, Jungtaek Kim, Adam Kosiorek, Seungjin Choi, and Yee~Whye
  Teh.
\newblock Set transformer: A framework for attention-based
  permutation-invariant neural networks.
\newblock In {\em International Conference on Machine Learning}, pages
  3744--3753. PMLR, 2019.

\bibitem[\protect\citeauthoryear{Luong \bgroup \em et al.\egroup
  }{2015}]{luong2015effective}
Thang Luong, Hieu Pham, and Christopher~D Manning.
\newblock Effective approaches to attention-based neural machine translation.
\newblock In {\em EMNLP}, 2015.

\bibitem[\protect\citeauthoryear{Paszke \bgroup \em et al.\egroup
  }{2019}]{paszke2019pytorch}
Adam Paszke, Sam Gross, Francisco Massa, Adam Lerer, James Bradbury, Gregory
  Chanan, Trevor Killeen, Zeming Lin, Natalia Gimelshein, Luca Antiga, et~al.
\newblock Pytorch: An imperative style, high-performance deep learning library.
\newblock {\em Advances in neural information processing systems},
  32:8026--8037, 2019.

\bibitem[\protect\citeauthoryear{Shi \bgroup \em et al.\egroup
  }{2019}]{shi2019variational}
Yuge Shi, Brooks Paige, Philip Torr, et~al.
\newblock Variational mixture-of-experts autoencoders for multi-modal deep
  generative models.
\newblock {\em Advances in Neural Information Processing Systems},
  32:15718--15729, 2019.

\bibitem[\protect\citeauthoryear{Suzuki \bgroup \em et al.\egroup
  }{2016}]{suzuki2016joint}
Masahiro Suzuki, Kotaro Nakayama, and Yutaka Matsuo.
\newblock Joint multimodal learning with deep generative models.
\newblock {\em arXiv preprint arXiv:1611.01891}, 2016.

\bibitem[\protect\citeauthoryear{Tan \bgroup \em et al.\egroup
  }{2019}]{tan2019envdrop}
Hao Tan, Licheng Yu, and Mohit Bansal.
\newblock Learning to navigate unseen environments: Back translation with
  environmental dropout.
\newblock In {\em Proceedings of the 2019 Conference of the North American
  Chapter of the Association for Computational Linguistics: Human Language
  Technologies, Volume 1 (Long and Short Papers)}, pages 2610--2621, 2019.

\bibitem[\protect\citeauthoryear{Tucker \bgroup \em et al.\egroup
  }{2018}]{tucker2018doubly}
George Tucker, Dieterich Lawson, Shixiang Gu, and Chris~J Maddison.
\newblock Doubly reparameterized gradient estimators for monte carlo
  objectives.
\newblock In {\em International Conference on Learning Representations}, 2018.

\bibitem[\protect\citeauthoryear{Wang \bgroup \em et al.\egroup
  }{2019}]{wang2019reinforced}
Xin Wang, Qiuyuan Huang, Asli Celikyilmaz, Jianfeng Gao, Dinghan Shen,
  Yuan-Fang Wang, William~Yang Wang, and Lei Zhang.
\newblock Reinforced cross-modal matching and self-supervised imitation
  learning for vision-language navigation.
\newblock In {\em Proceedings of the IEEE/CVF Conference on Computer Vision and
  Pattern Recognition}, pages 6629--6638, 2019.

\bibitem[\protect\citeauthoryear{Wu and Goodman}{2018}]{wu2018multimodal}
Mike Wu and Noah Goodman.
\newblock Multimodal generative models for scalable weakly-supervised learning.
\newblock In {\em Proceedings of the 32nd International Conference on Neural
  Information Processing Systems}, pages 5580--5590, 2018.

\bibitem[\protect\citeauthoryear{Yu \bgroup \em et al.\egroup
  }{2020}]{yu2020take}
Felix Yu, Zhiwei Deng, Karthik Narasimhan, and Olga Russakovsky.
\newblock Take the scenic route: Improving generalization in
  vision-and-language navigation.
\newblock In {\em Proceedings of the IEEE/CVF Conference on Computer Vision and
  Pattern Recognition Workshops}, pages 920--921, 2020.

\bibitem[\protect\citeauthoryear{Zhao \bgroup \em et al.\egroup
  }{2019}]{zhao2017infovae}
Shengjia Zhao, Jiaming Song, and Stefano Ermon.
\newblock Infovae: Balancing learning and inference in variational
  autoencoders.
\newblock In {\em The Thirty-Third {AAAI} Conference on Artificial
  Intelligence}, pages 5885--5892, 2019.

\end{thebibliography}

\end{document}